\documentclass{article}

\usepackage[final]{neurips_2025_ml4ps}

\usepackage[utf8]{inputenc} % allow utf-8 input
\usepackage[T1]{fontenc}    % use 8-bit T1 fonts
\usepackage{hyperref}       % hyperlinks
\usepackage{url}            % simple URL typesetting
\usepackage{booktabs}       % professional-quality tables
\usepackage{amsfonts}       % blackboard math symbols
\usepackage{nicefrac}       % compact symbols for 1/2, etc.
\usepackage{microtype}      % microtypography
\usepackage{xcolor}         % colors
\setcitestyle{square,sort,comma,numbers}
\usepackage{graphicx}
\usepackage{caption}
\usepackage{subcaption}
\usepackage{algorithm}
\usepackage{algorithmic}
\newtheorem{theorem}{Theorem}[section]
\newtheorem{corollary}{Corollary}[theorem]
\usepackage{amsmath,amsfonts,bm}

\def\eqref#1{equation~\ref{#1}}
\def\1{\bm{1}}

\def\rvv{{\mathbf{v}}}
\def\rvw{{\mathbf{w}}}
\def\rvx{{\mathbf{x}}}

\DeclareMathAlphabet{\mathsfit}{\encodingdefault}{\sfdefault}{m}{sl}
\SetMathAlphabet{\mathsfit}{bold}{\encodingdefault}{\sfdefault}{bx}{n}

\def\gM{{\mathcal{M}}}
\def\gN{{\mathcal{N}}}

\def\sR{{\mathbb{R}}}

\title{Intrinsic Dimension Estimation for Radio Galaxy Zoo using Diffusion Models 
}

\author{%
  Joan Font-Quer Roset\\
  Department of Computer Science\\
  The University of Manchester\\
  Manchester, United Kingdom \\
  \texttt{joan.font-quer@postgrad.manchester.ac.uk} \\
  \and
  \textbf{Devina Mohan}\\
  Department of Physics and Astronomy\\
  The University of Manchester\\
  Manchester, United Kingdom \\
\texttt{devina.mohan@postgrad.manchester.ac.uk}\\
  \and
  \textbf{Anna Scaife}\thanks{The Alan Turing Institute, 96 Euston Rd, London, UK \texttt{a.scaife@turing.ac.uk}}\\
  Department of Physics and Astronomy\\
  The University of Manchester\\
  Manchester, United Kingdom \\
  \texttt{anna.scaife@manchester.ac.uk}
}

\begin{document}
\maketitle
\begin{abstract}
In this work, we estimate the intrinsic dimension (iD) of the Radio Galaxy Zoo (RGZ) dataset using a score-based diffusion model. We examine how the iD estimates vary as a function of Bayesian neural network (BNN) energy scores, which measure how similar the radio sources are to the MiraBest subset of the RGZ dataset. We find that out-of-distribution sources exhibit higher iD values, and that the overall iD for RGZ exceeds those typically reported for natural image datasets. Furthermore, we analyse how iD varies across Fanaroff-Riley (FR) morphological classes and as a function of the signal-to-noise ratio (SNR). While no relationship is found between FR I and FR II classes, a weak trend toward higher SNR at lower iD. Future work using the RGZ dataset could make use of the relationship between iD and energy scores to quantitatively study and improve the representations learned by various self-supervised learning algorithms.
\end{abstract}

\section{Introduction}

The field of astronomy has experienced a deluge of data from large-scale surveys from new observational facilities such as Euclid \citep{mellier2024euclid} and the James Webb Space Telescope \citep[JWST;][]{pontoppidan2022jwst}. This has led to the increased adoption of deep learning models to classify or analyse data \citep{lines2025revolution}. However, these models are constrained by the scarcity of labelled data. This has motivated the development of techniques that can analyse large unlabelled datasets, such as the Radio Galaxy Zoo \cite[RGZ;][]{wong2025radio}. 

According to the manifold hypothesis, high-dimensional data often lies in low dimensional submanifolds \cite{bengio2013representation}, whose intrinsic dimensions (iD) reflects the effective number of degrees of freedom in a data distribution. Estimating iD then provides a useful tool to not only analyse data, but also to analyse models trained on the data.

Recent work has shown that diffusion models encode the iD of data manifolds \cite{stanczuk2024diffusion} through their score functions. We use this method to estimate the intrinsic dimension of the RGZ dataset using a score-based diffusion model, and compare the resulting estimates against traditional statistical methods (LPCA \cite{PCA}\cite{LPCA}, PPCA \cite{PPCA}, and MLE \cite{MLE}). 

We relate the iD estimates to Bayesian neural network (BNN) energy scores obtained from a Hamiltonian Monte Carlo BNN trained on the MiraBest dataset, which is a labelled subset of the RGZ data. These energy scores provide a measure of how in or out of distribution a radio source is relative to the labelled benchmark \cite{pmlr-v244-mohan24a}. Furthermore, we examine how the estimated iD differs across Fanaroff-Riley (FR) morphological classes, using existing labels from recent work on foundation model based radio galaxy classification \cite{FRRGZClass}. The FR classification divides radio galaxies into two morphological types:
FR I sources show peak brightness near the core and fade with distance, FR II sources exhibit bright outer lobes separated by a fainter core \cite{fr1974}.

\section{Estimating intrinsic dimension of data with diffusion models}
\label{sec:diffusion_id}
% \cite{stanczuk2024diffusion} show that score diffusion models encode the intrinsic dimension of data by approximating the score function for small perturbations 

In a diffusion model, the data distribution $p_0$ is perturbed using a series of noise-corrupted distributions $p_t$ using a stochastic differential equation \cite{song2020score}. Diffusion models are trained to approximate the score vector $\nabla_{\rvx} \ln(p_t \rvx)$ with a neural network $s_\theta(x_t, t)$. Once the score function is approximated, it can be used to generate new samples from the data distribution. The recently proposed method provides a theoretical foundation, showing that the approximate local dimensionality of the data manifold can be recovered from the score vector \citep{stanczuk2024diffusion}. In experiments with the MNIST dataset, it is found that digits with different geometric complexity have different iDs. For example the digit $1$ has iD $= 66$ and the digit $9$ has iD $= 152 $. Previous work by \citep{pope2021the} also estimated the iD of several image datasets using a maximum likelihood estimation (MLE) based method, but later work showed that this severely underestimates iD.

\subsection{Theoretical Foundation}

\begin{algorithm}
\caption{Estimate the Intrinsic Dimension at $\rvx_0$ \cite{stanczuk2024diffusion}}
\label{alg:intrinsic-dim}
\begin{algorithmic}
\REQUIRE $s_\theta$ -- trained diffusion model (score),\\
\hspace{3em} $t_0$ -- sampling time,\\
\hspace{3em} $K$ -- number of score vectors.
\STATE Sample $\rvx_0 \sim p_0(\rvx)$ from the dataset
\STATE $d \leftarrow \dim(\rvx_0)$
\STATE $S \leftarrow \text{empty matrix}$
\FOR{$i = 1, \dots, K$}
    \STATE Sample $\rvx^{(i)}_{t_0} \sim \mathcal{N}(\rvx_0, \sigma_{t_0}^2 I)$
    \STATE Append $s_\theta(\rvx^{(i)}_{t_0}, t_0)$ as a new column to $S$
\ENDFOR
\STATE $(s_i)_{i=1}^d, (\rvv_i)_{i=1}^d, (\rvw_i)_{i=1}^d \leftarrow \mathrm{SVD}(S)$
\STATE $\hat{k}(\rvx_0) \leftarrow d - \arg\max_{i=1,\ldots,d-1}(s_i - s_{i+1})$
\RETURN $\hat{k}(\rvx_0)$
$(s_i)_{i=1}^d, (\rvv_i)_{i=1}^d, (\rvw_i)_{i=1}^d$ denote singular values, left and right singular vectors respectively.
\end{algorithmic}
\end{algorithm}

\begin{theorem} \label{th:main_res}
    For any point $\rvx \in \sR^d$ sufficiently close to a compact embedded manifold $\gM$, and a sufficiently small diffusion time $t$, the score vector $\nabla_{\rvx} \ln(p_t (\rvx))$ points directly at the projection of $\rvx$ on the manifold.
\end{theorem}
\begin{corollary} \label{cor:zero_ratio}
    The ratio of the projection of the score vector $\nabla_{\rvx} \ln(p_t \rvx)$ on the tangent space of the manifold $T_{\pi(\rvx) \gM}$ to the normal space of the manifold $\gN_{\pi(\rvx) \gM}$ approaches 0 as $t$ approaches 0.
\end{corollary}
%
% It is important to note that 
Theorem~\ref{th:main_res} only works under certain assumptions:
(i) The data lies on a compact, smooth, embedded manifold;
(ii) $p_0$ has a smooth (differentiable) density function defined over the manifold; and
(iii) the density function is strictly positive on the manifold (i.e. there are no zero-probability regions).
Theorem~\ref{th:main_res} and Corollary~\ref{cor:zero_ratio} underpin the method's central idea: the score vectors of perturbed points near the manifold will concentrate on the normal space to the manifold at $\rvx_0$ as $t \rightarrow 0$. Therefore, its projection to tangent space vanishes as $t \rightarrow 0$, and so the number of vanishing singular values corresponds to the dimension of the tangent space. The iD is then the ambient dimension minus the rank of the normal space. Algorithm~\ref{alg:intrinsic-dim} describes the method proposed in the paper for estimating the iD at a point $\rvx_0$ using a trained diffusion model. For each $i \in \{ 1, \dots K\}$, a sample $\rvx^{(i)}_{t_0}$ is drawn from the forward process transition kernel $p_0 (\rvx^{(i)}_{t_0} | \rvx_0) = \mathcal{N}(\rvx_0, \sigma_{t_0}^2 I)$, and its score vector $s_\theta (\rvx^{(i)}_{t_0})$ is computed. Each of these score vectors forms a column of a $d \times K$ matrix $S$, where $d$ is the ambient dimension of $\rvx_0$. Singular value decomposition (SVD) is applied to the matrix $S$, and the iD is estimated as $d - i^\ast$, where $i^\ast$ corresponds to the index of the largest gap between singular values $\arg\max_{i=1,\ldots,d-1}(s_i - s_{i+1})$.

\section{Experimental Setup}
\textbf{Data}
We use the Radio Galaxy Zoo Data Release 1 \citep{wong2025radio} as our unlabelled dataset of radio sources to train the diffusion model. Details of the MiraBest dataset \cite{Porter2020MiraBestDataset} which is used to train the BNN is provided in Appendix \ref{ref: mb}.

\textbf{Radio Galaxy Zoo Data Release 1}
The Radio Galaxy Zoo (RGZ) is a citizen science project which contains extended radio sources from the FIRST and ATLAS radio surveys cross-matched with their host galaxies in the infrared from the WISE and SpitzerSpace Telescope, 
%where possible
respectively \citep{wong2025radio}. Instead of classifying morphologies based on the FR scheme, RGZ classifies sources into the number of components, which are discrete patches of emission enclosed by a contour of constant brightness, and the number of peaks of maximum brightness within a component. The catalogue provides $100,185$ classifications along with a weighted user consensus level. The user consensus level is calibrated by comparing each individual users' classifications to that of an expert for a smaller sample of $20$ sources. 
% In total expert classifications were available for $1000$ sources. 
We use RGZ Data Release 1 in this work which contains classifications with a user consensus level greater than or equal to  $0.65$. FIRST-based catalogues make up $99.4 \%$ of DR1. We select radio sources with angular size $> 20$ arcsecond. The images are of size 72x72.
% , since this cutoff was also applied during pre-training of the BYOL model. 
% Removed duplicates which had the same rgz id in the dataset.

\begin{figure}[t] %{\textwidth}
    \centering
    \includegraphics[width=1.0\textwidth]{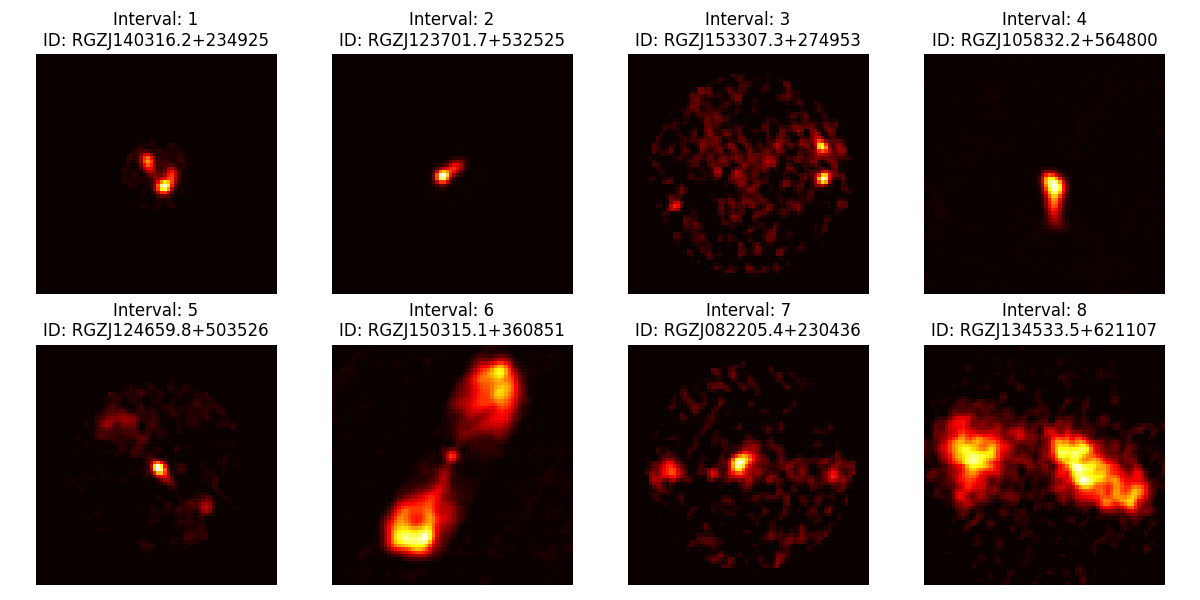}
    \caption{Example images from the Radio Galaxy Zoo dataset taken from different intervals of the mean energy distribution. 
    % Galaxies from different intervals are increasingly distant from the canonical radio galaxy
    }
    \label{fig:rgz_mean}
\end{figure}

\textbf{Bayesian neural network}
To impose some structure in our analysis of the iD estimates of RGZ, we create subsets of the dataset using a BNN. We use the BNN based on Hamiltonian Monte Carlo (HMC) from the benchmark
provided for radio galaxy classification in \citep{pmlr-v244-mohan24a}. This BNN is trained using the MiraBest dataset to classify radio galaxies into FRI and FRII and performs the best in terms of being able to clearly detect distribution shifted radio galaxies based on energy scores \citep{liu2020energy}. 
% BNN training details are provided in the appendix.  
We pass the RGZ dataset through the trained BNN and obtain a distribution of the scalar energy scores, $\mathrm{E}(x; f):= -T . \,  \textrm{log} \sum_i^K e^{f_{i} (x) / T }$,
%
% \begin{equation}
%     \mathrm{E}(x; f) = -T . \,  \textrm{log} \sum_i^K e^{f_{i} (x) / T } ,
% \end{equation}
%
for all the sources in the dataset, $x$, using the logit values, $f_i(x)$, for each class, $i$. The temperature term, $T$, is set to $1$. We use $N = 200$ posterior samples from HMC to construct the energy score distributions and calculate the mean and standard deviation of the distribution for each RGZ source. We then fit a log-normal distribution to the distribution of the mean and standard deviation estimates of the entire RGZ dataset and calculate an interval label for each source based on how many standard deviations away a source is from the mean of the mean energy values and the mean of the standard deviation values of the energy distributions. We find that the RGZ sources fall into 8 mean intervals and 6 standard deviation intervals. 
% We use these intervals as a measure of increasingly distribution sifted samples. 
Examples of galaxies from each mean and standard deviation interval are shown in Figure \ref{fig:rgz_mean} and Figure \ref{fig:rgz_std}, respectively. 

\textbf{Diffusion Intrinsic Dimension Estimation}

We train a score-based diffusion model on the RGZ data using the weighted denoising score matching objective \cite{song2020score} and a Denoising diffusion probabilistic model (DDPM) architecture \cite{ho2020denoising}. The diffusion model is trained for 400 epochs, selecting the model with the lowest validation loss.

We use the method described in Section \ref{sec:diffusion_id} to estimate the intrinsic dimension of data from different intervals of the mean and standard deviation of energy values. We perturb all the samples $k=328$ times to calculate the score vector for each noisy point using the diffusion model. The score vectors are collected into a matrix and SVD is performed to estimate iD. The score spectrum plots are shown in Figure \ref{fig:svd}.

\textbf{Baseline Intrinsic Dimension Estimation}

We estimate the iD for each sample in the RGZ dataset using the MLE method with two neighbouring settings: $m=5$ and $m=20$. In addition, we compute the per-sample estimates using local PCA via the Fukunaga-Olsen method \cite{Fukunaga1971}, setting alpha to $0.05$. Probabilistic PCA produces only global estimates, so iD values were instead computed per label.

\begin{figure}
     \centering
     \begin{subfigure}[b]{0.49\textwidth}
         \centering
         \includegraphics[width=\textwidth]{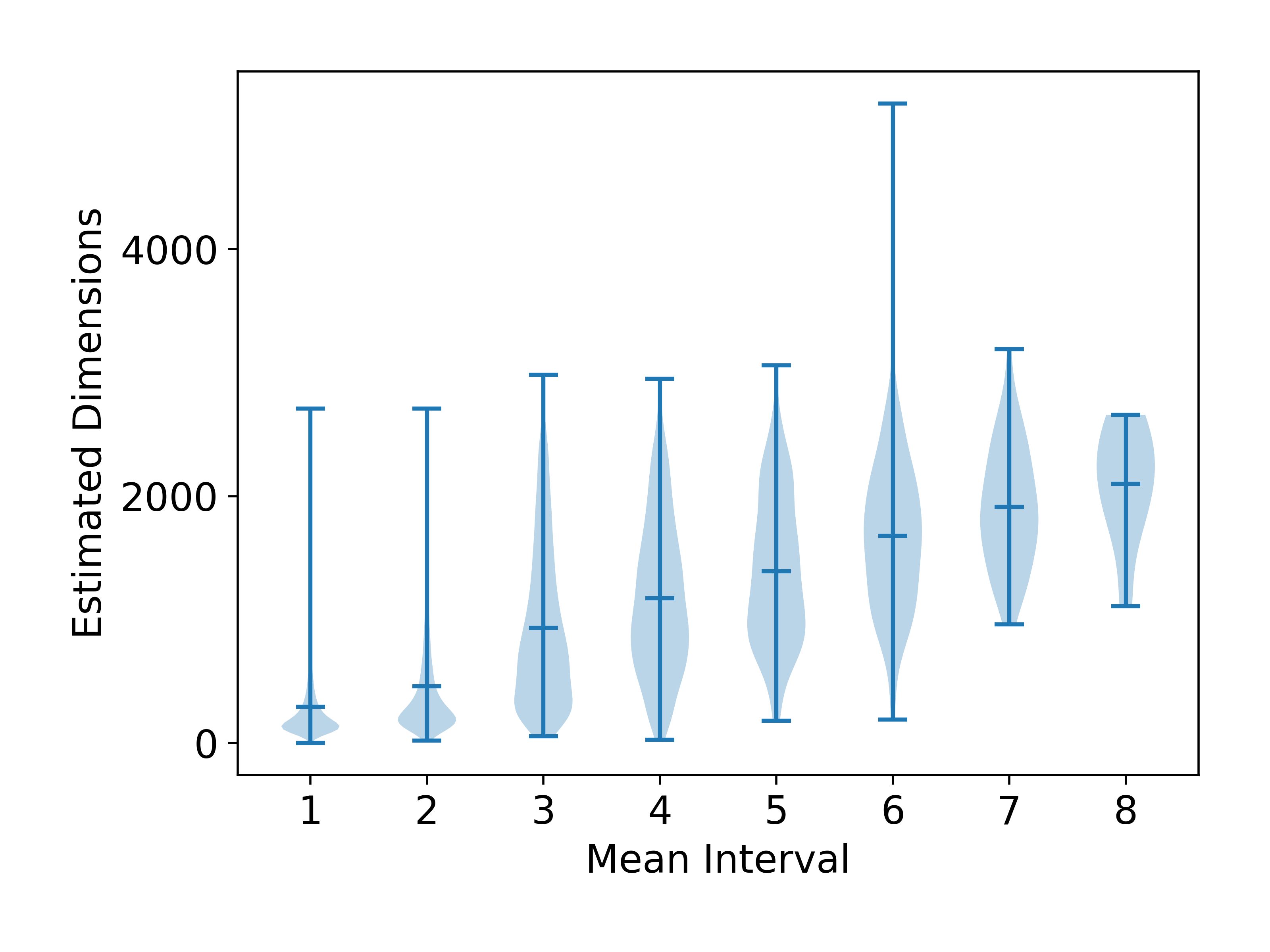}
         \caption{}
         \label{fig:id_mean}
     \end{subfigure}
     \hfill
     \begin{subfigure}[b]{0.49\textwidth}
         \centering
         \includegraphics[width=\textwidth]{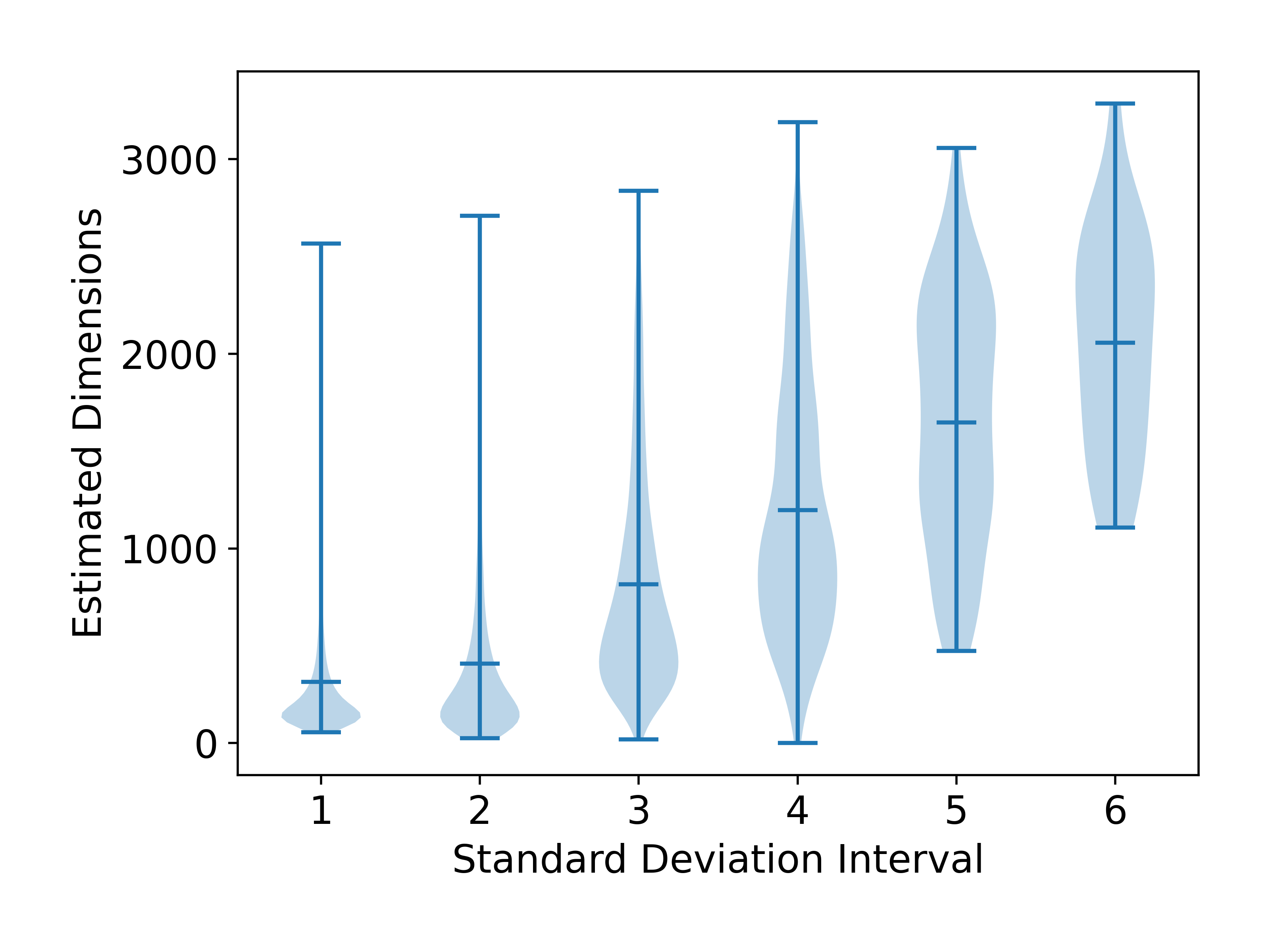}
         \caption{}
         \label{fig:id_std}
     \end{subfigure}

        \caption{Intrinsic dimension estimates for the RGZ dataset as a function of the (a) mean interval and (b) standard deviation interval of the energy distribution from the Hamiltonian Monte Carlo based Bayesian neural network trained on the MiraBest dataset.}
        \label{fig:rgz_id}
\end{figure}

\begin{figure}
    \centering
    \includegraphics[width=0.5\linewidth]{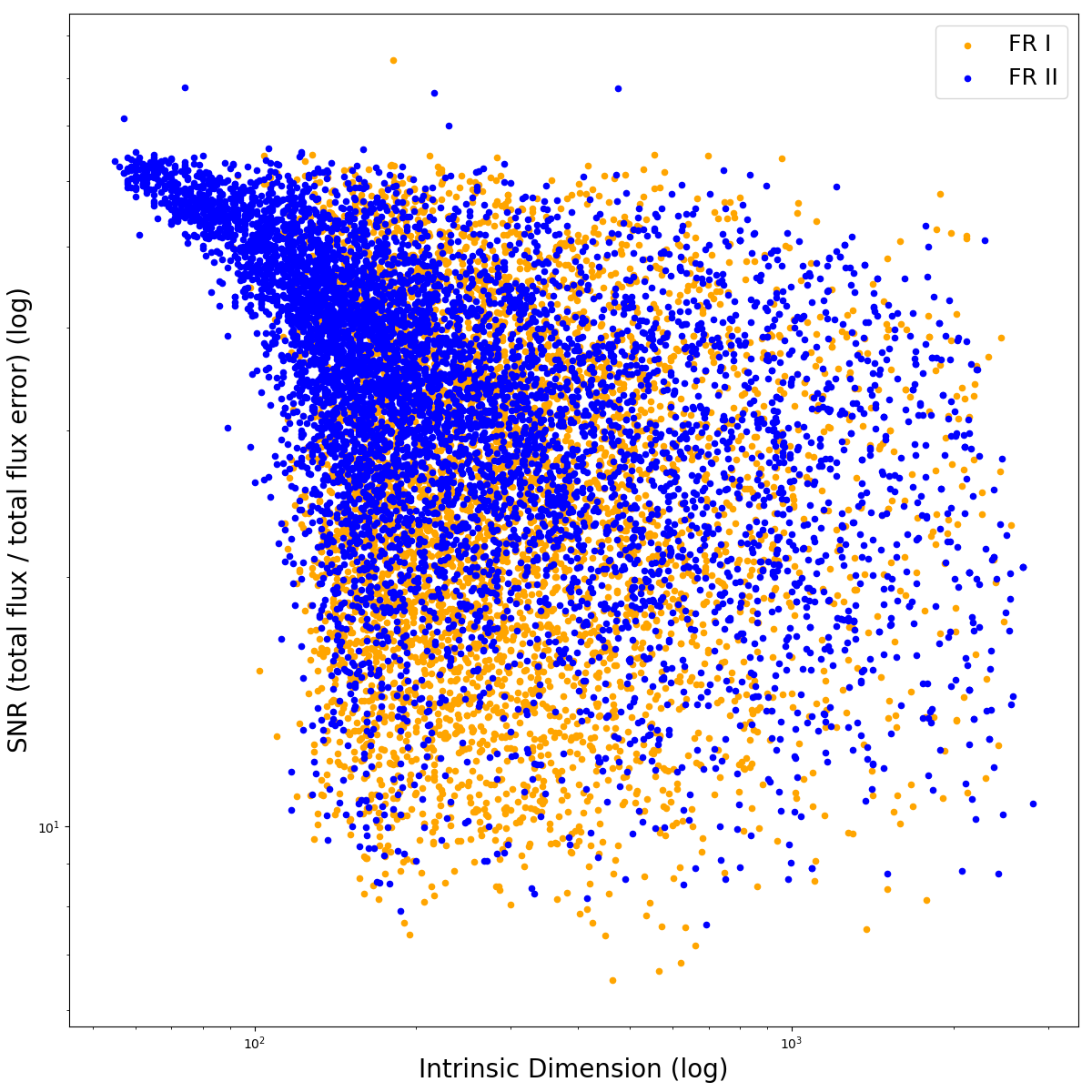}
    \caption{Signal to noise ratio versus the intrinsic dimension for a subset of galaxies labelled by Fanaroff-Riley class. Both axes are shown in logarithmic scale.}
    \label{fig:snr_vs_id}
\end{figure}

\begin{table}[t]
\centering
\caption{Standard deviation interval estimates across benchmarks. ``N/A'' indicates the method was not applicable due to insufficient samples.}
\label{tab:std-intervals}
\begin{tabular}{lcccccc}
\toprule
Method       & 1      & 2      & 3      & 4      & 5      & 6      \\
\midrule
MLE ($m=5$)  & 9.318  & 9.172  & 17.546 & 19.157 & 18.394 & 20.681 \\
MLE ($m=20$) & 8.182  & 7.954  & 15.667 & 17.207 & 18.277 & 20.353 \\
Local PCA    & 10.571 & 11.892 & 16.907 & 19.144 & 20.586 & 21.556 \\
PPCA         & 4977   & 3777   & N/A    & N/A    & N/A    & N/A    \\
\bottomrule
\end{tabular}
\end{table}

\begin{table}[t]
\centering
\caption{Mean interval estimates across benchmarks. ``N/A'' indicates the method was not applicable due to insufficient samples.}
\label{tab:mean-intervals}
\begin{tabular}{lcccccccc}
\toprule
Method        & 1      & 2      & 3      & 4      & 5      & 6      & 7      & 8      \\
\midrule
MLE ($m=5$)   & 9.249  & 9.927  & 12.261 & 13.437 & 17.958 & 21.655 & 26.092 & 40.034 \\
MLE ($m=20$)  & 8.086  & 8.645  & 11.309 & 12.728 & 16.174 & 21.155 & 25.298 & 38.033 \\
Local PCA     & 10.702 & 12.064 & 15.368 & 16.653 & 17.331 & 17.523 & 16.725 & 17.889 \\
PPCA          & 4357   & 3638   & N/A    & N/A    & N/A    & N/A    & N/A    & N/A    \\
\bottomrule
\end{tabular}
\end{table}

\section{Results \& Conclusion}

Figure \ref{fig:rgz_id} and Tables \ref{tab:mean-intervals}-\ref{tab:std-intervals} show that the estimated iD increase with interval number for both the mean and standard deviations. This suggests that the object identified with the BNN as out-of-distribution have a higher iD than the objects that are in distribution.

Comparisons with the classical estimators (MLE, LPCA, and PPCA) show that, while they all broadly follow the same trend, the diffusion estimates are much larger in magnitude. This could be due to the fact that the classical methods tend to underestimate the iD of high dimensional data \cite{underest2023}. In addition, the diffusion iD estimates are also much larger compared to those reported for natural image datasets in previous work \cite{pope2021the, stanczuk2024diffusion}. This could be due to the inherently noisy nature of radio astronomy data. 

Figure \ref{fig:snr_vs_id} shows the relationship between the signal-to-noise ratio (SNR) and the iD of each galaxy, along with their FR morphology. Overall, there is no strong relationship between SNR and iD. However, at low iD  values ($<100$), FR II sources tend to exhibit higher SNRs, suggesting that galaxies with lower iD are less affected by noise. On the other hand, if a galaxy has a high SNR, it does not necessarily have a low iD.

In future work, we will extend this approach to estimate the iD using the features learned by self-supervised learning based models trained on the RGZ dataset (\cite{slijepcevic2024radio}). We will use the results to determine what degree of compression can be achieved for different subsets of the data.

% We argue that we can use a well-calibrated probability mass 
% \section{Conclusion}
% In future work, intrinsic dimension of the features learned by self-supervised learning models trained on RGZ can be examined to determine what degree of 

%summarize position paper
% \cite{li2025position} argue that using supervised models trained on some in-distribution data is a misspecified objective for detecting out-of-distribution data based on uncertainty or features. For instance, predictive uncertainty is measured over the output labels and high uncertainty over model outputs should not be conflated with the presence of out-of-distribution data. 
% "we are interested in typicality rather than density but different notions of typicality lead to very different OoD behaviour." We propose to measure typicality using the energy scores from a Bayesian neural network trained on a small labelled dataset. 

% This method will only be as good as the posterior approximation.
\begin{ack}
This work was supported by the Engineering and Physical Sciences Research Council EP/Y030826/1.
AMS gratefully acknowledges support from an Alan Turing Institute AI Fellowship EP/V030302/1.
This work was done under the supervision of Anna Scaife, Julia Handl, and Mingfei Sun.
\end{ack}

\bibliography{references}

\clearpage

\appendix

\section{MiraBest dataset}
\label{ref: mb}
\textbf{MiraBest Dataset}
The MiraBest dataset used in this work consists of 1256 images of radio galaxies of $150\times 150$ pixels pre-processed to be used specifically for deep learning tasks \citep{porter2023mirabest}. The dataset was constructed using the sample selection and classification described in \cite{Miraghaei2017TheEnvironment}, who made use of the parent galaxy sample from \cite{Best2012OnProperties}. Optical data from data release 7 of Sloan Digital Sky Survey \citep[SDSS DR7;][]{abazajian2009} was cross-matched with NRAO VLA Sky Survey  \citep[NVSS;][]{condon1998} and Faint Images of the Radio Sky at Twenty-Centimeters  \citep[FIRST;][]{becker1995} radio surveys. 
The galaxies are labelled using the FRI and FRII morphological types based on the definition of \citep{fanaroff1974morphology} and further divided into their subtypes. In addition to labelling the sources as FRI, FRII and their subtypes, each source is also flagged as `Confident' or `Uncertain' to indicate the human classifiers' confidence while labelling the dataset. In this work we use the MiraBest Confident subset and consider only the binary FRI/FRII classification. The training and validation sets are created by splitting the predefined training data into a ratio of 80:20. The final split consists of 584 training samples, 145 validation samples, and 104 withheld test samples. 
% \section{Additional plots}
% Figure \ref{fig:rgz_std}
\begin{figure}[t] %{\textwidth}
    \centering
    \includegraphics[width=1.0\textwidth]{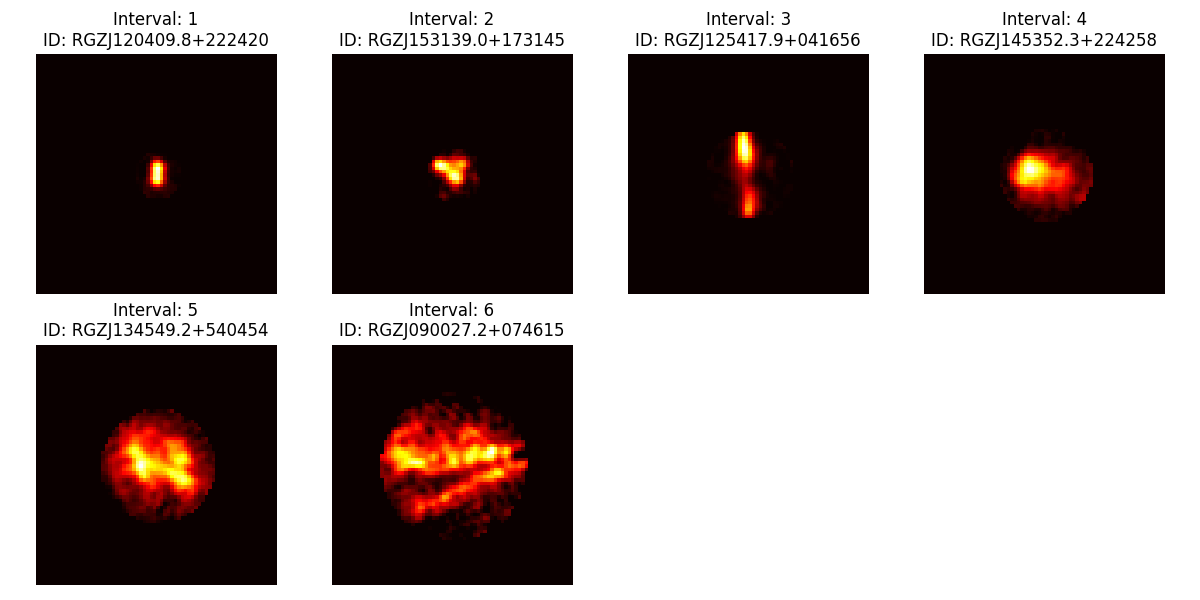}
    \caption{Images from the Radio Galaxy Zoo dataset from different intervals of the standard deviation of energy scores.}
    \label{fig:rgz_std}
\end{figure}

\section{Score Spectra}

\begin{figure}
     \centering
     \begin{subfigure}[b]{0.49\textwidth}
         \centering
         \includegraphics[width=\textwidth]{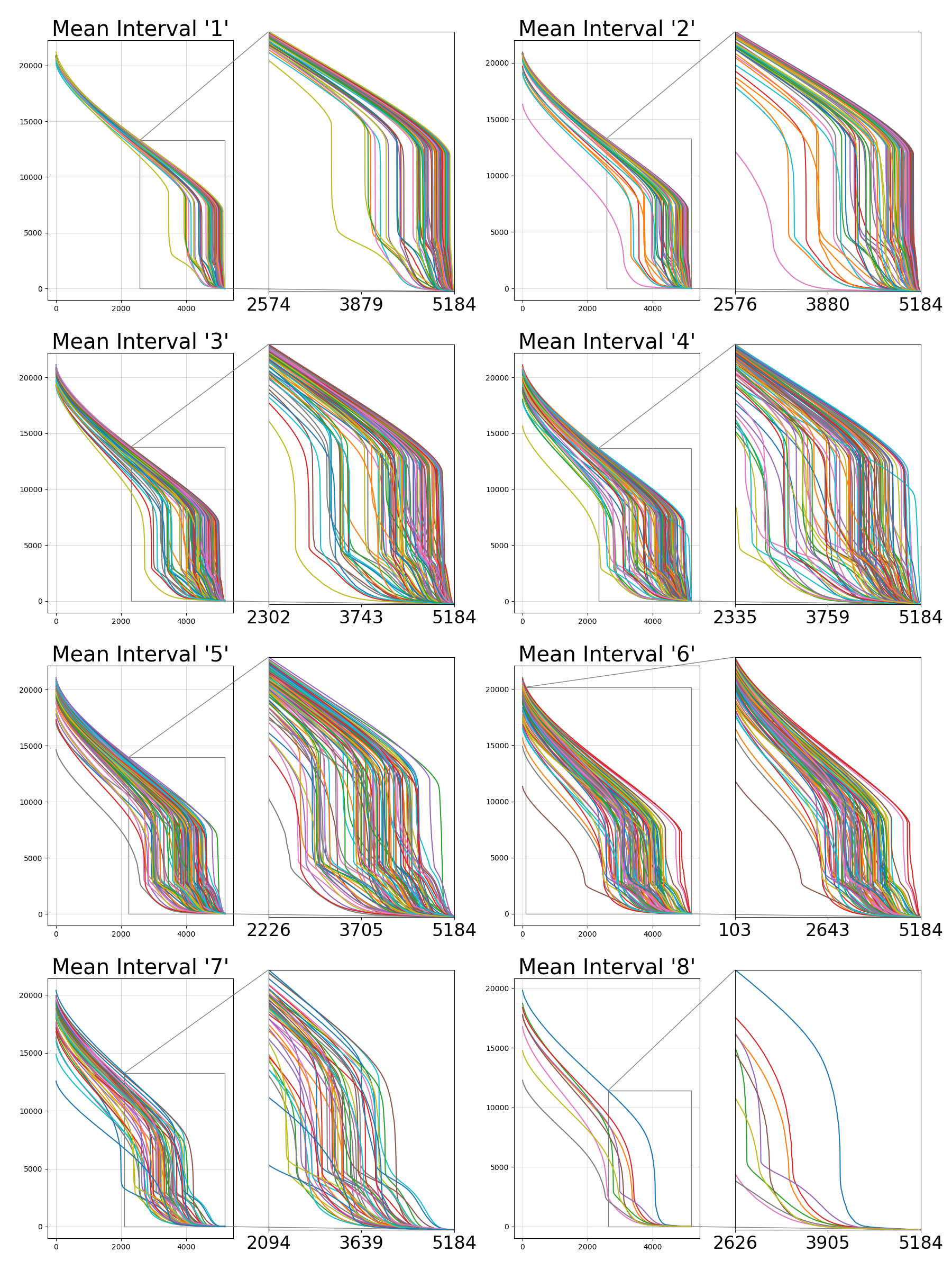}
         \caption{}
         \label{fig:svd_mean}
     \end{subfigure}
     \hfill
     \begin{subfigure}[b]{0.49\textwidth}
         \centering
         \includegraphics[width=\textwidth]{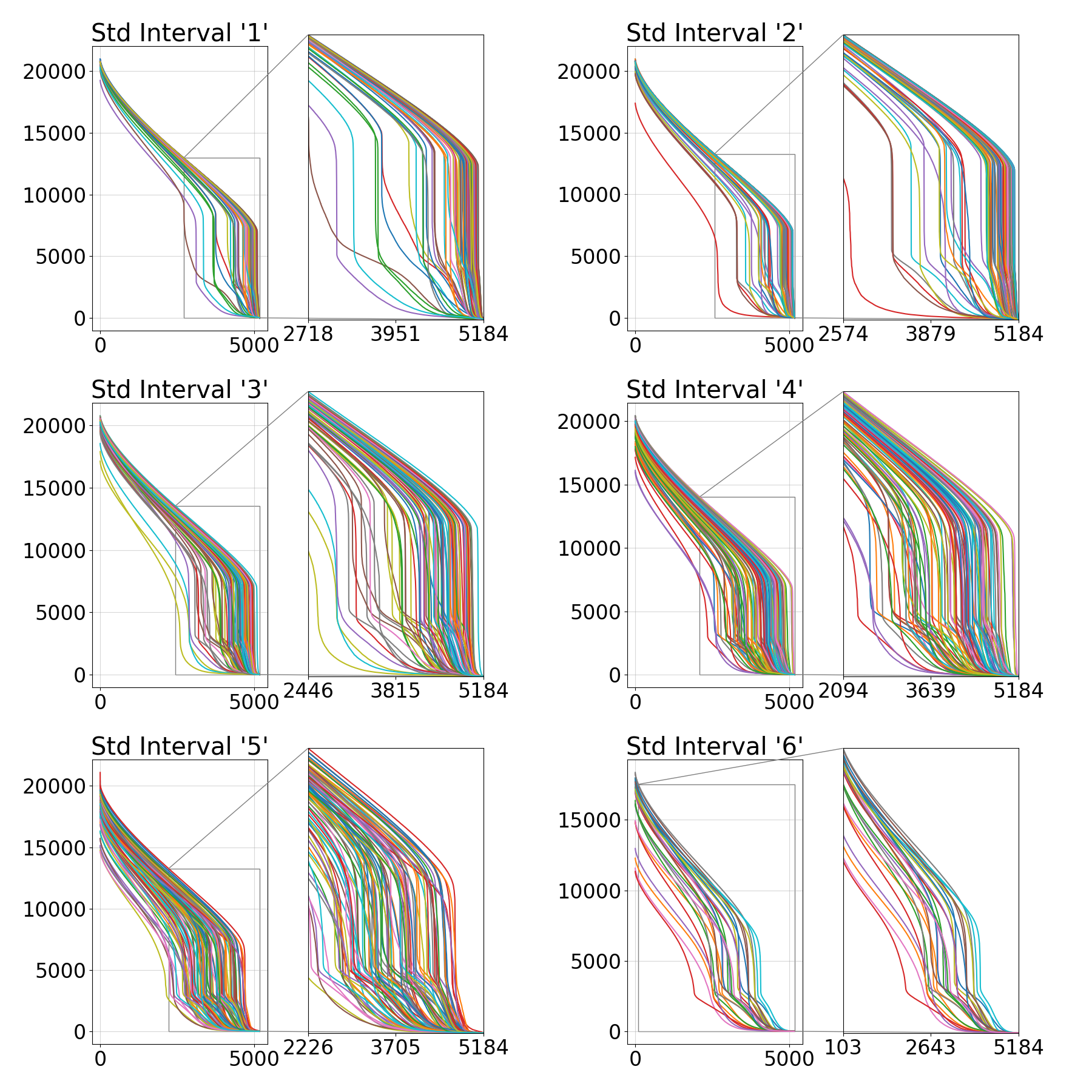}
         \caption{}
         \label{fig:svd_std}
     \end{subfigure}

        \caption{Score spectrum plots for up to 100 randomly selected RGZ sources from different (a) mean and (b) standard deviation intervals of the energy distribution. Here, the x-axis shows the singular values which go up to the ambient dimension (72x72 for RGZ images) and the y-axis shows the magnitude of each singular value. The point at which there is a sharp drop in singular values indicates the normal dimension. This can be subtracted from the ambient dimension to calculate the intrinsic dimension.}
        \label{fig:svd}
\end{figure}

Figure \ref{fig:svd} shows the score spectra for $100$ randomly selected samples for each label.

\end{document}